\documentclass[lettersize,journal]{IEEEtran}
\usepackage{cite}
\usepackage{xcolor}
\usepackage{amsmath,amssymb,amsfonts}
\usepackage{algorithmic} 
\usepackage{graphicx}
\usepackage{textcomp}
\def\BibTeX{{\rm B\kern-.05em{\sc i\kern-.025em b}\kern-.08em
    T\kern-.1667em\lower.7ex\hbox{E}\kern-.125emX}}

\usepackage{xtab}
\usepackage{graphicx}
\usepackage{array} 
\usepackage{multirow} 
\usepackage[table,xcdraw]{xcolor} 
\usepackage{placeins} 
\usepackage{booktabs}    
\usepackage{ragged2e}
\usepackage{makecell}    
\usepackage{array}   
\usepackage[protrusion=true,expansion=true]{microtype}
\pdfoutput=1
\usepackage{tabularx}
\usepackage{url}
\usepackage[breaklinks=true]{hyperref}

\usepackage{placeins}  
\usepackage{float}     

\definecolor{Gray}{gray}{0.9}

\definecolor{RedMod1}{RGB}{26, 38, 87} 
\definecolor{RedMod2}{RGB}{26, 38, 87} 
\definecolor{RedMod}{RGB}{26, 38, 87} 

\definecolor{darkyellow}{RGB}{159, 174, 191} 
\definecolor{salmon}{RGB}{208, 222, 196}
\definecolor{darksalmon}{RGB}{171, 196, 150}
\definecolor{darkgreen}{RGB}{180, 230, 198}
\definecolor{lightgreen}{RGB}{209, 233, 201}

\definecolor{lightred}{RGB}{255,211,222}
\definecolor{darkred}{RGB}{252,150,167}
 \definecolor{LightBlue}{rgb}{0.8,0.89,1} 
\definecolor{DarkBlue}{rgb}{0.57,0.71,0.82} 
\definecolor{MagLight}{rgb}{1, 0.89, 0.8}
\usepackage{enumitem}

\usepackage[most]{tcolorbox}
  \makeatletter
\newcommand\notsotiny{\@setfontsize\notsotiny\@vipt\@viipt}
\makeatother
\begin{document}

\title{Multi-Modal Multi-Task (M3T) Federated Foundation Models for Embodied AI: Potentials and Challenges for Edge Integration}

\author{Kasra Borazjani,~\IEEEmembership{Student Member,~IEEE,} Payam Abdisarabshali,~\IEEEmembership{Student Member,~IEEE,} Fardis Nadimi, \\ \IEEEmembership{Student Member,~IEEE,} Naji Khosravan,~\IEEEmembership{Member,~IEEE,} Minghui Liwang,~\IEEEmembership{Senior Member,~IEEE,} \\ Xianbin Wang,~\IEEEmembership{Fellow,~IEEE,} Yiguang Hong,~\IEEEmembership{Fellow,~IEEE,} Seyyedali Hosseinalipour,~\IEEEmembership{Senior Member,~IEEE}
\vspace{-0mm}
}



\maketitle

\begin{abstract}
As embodied AI systems become increasingly multi-modal, personalized, and interactive, they must learn effectively from diverse sensory inputs, adapt continually to user preferences, and operate safely under resource and privacy constraints. 
These challenges expose a pressing need for machine learning models capable of swift, context-aware adaptation while balancing model generalization and personalization. 
Here, two methods emerge as suitable candidates, each offering parts of these capabilities: multi-modal multi-task foundation models (M3T-FMs) provide a pathway toward generalization across tasks and modalities, whereas federated learning (FL) offers the infrastructure for distributed, privacy-preserving model updates and user-level model personalization. However, when used in isolation, each of these approaches falls short of meeting the complex and diverse capability requirements of real-world embodied AI environments.
In this vision paper, we introduce \textit{multi-modal multi-task federated foundation models (M3T-FFMs) for embodied AI}, a new paradigm that unifies the strengths of M3T-FMs with the privacy-preserving distributed training nature of FL, enabling intelligent systems at the wireless edge.  We collect critical deployment dimensions of M3T-FFMs in embodied AI ecosystems under a unified framework, which we name \textbf{``EMBODY"}: \underline{E}mbodiment heterogeneity, \underline{M}odality richness and imbalance, \underline{B}andwidth and compute constraints, \underline{O}n-device continual learning, \underline{D}istributed control and autonomy, and \underline{Y}ielding safety, privacy, and personalization. For each, we identify concrete challenges and envision actionable research directions. We also present an evaluation framework for deploying M3T-FFMs in embodied AI systems, along with the associated trade-offs. Finally, we present a prototype implementation of M3T-FFMs and evaluate their energy and latency performance. To foster further research in this largely untapped area, we share our implementation through an open-source repository (GitHub: \url{https://github.com/payamsiabd/M3T-FFM-EmbodiedAI}).
\end{abstract}

\maketitle



\section{Introduction}
Embodied AI refers to artificial intelligence systems that are physically situated in the world --- typically within robots or agents that can sense, act, and learn through interaction with their environment, e.g., ``Figure 01" by Figure AI, ``Boston Dynamics' Spot", and ``Meta Quest" or ``Apple Vision Pro" extended reality (XR) devices \cite{duan2022survey,figai}. Embodied AI is not just redefining the role of intelligent systems, it is reimagining their very nature. 
What fundamentally distinguishes the next generation of embodied AI agents from traditional AI systems, such as Large Language Models (LLMs) or static vision classifiers, is their demand for \textit{interactive, physically-grounded intelligence}. In particular, embodied agents must continuously perceive the world through multiple \textit{sensor modalities} (e.g., vision, touch, audio), interact with \textit{dynamic environments}, and adapt to \textit{diverse tasks}, from object manipulation and social interaction to navigating unstructured terrain for search-and-rescue and assisting surgeons in hospitals.

These requirements cannot be met by the deployment of narrowly trained, single-task models at these AI agents. Instead, they naturally align with the capabilities of emerging \textit{multi-modal multi-task (M3T) foundation models (FMs)}, which are large-scale architectures \textit{often} pretrained on diverse datasets that span language, visual scenes, and human instructions~\cite{driess2023palm}. M3T-FMs can provide a unified semantic backbone for embodied agents, enabling them to interpret instructions, understand environments, and plan actions. For example, a kitchen robot could leverage a single M3T-FM to recognize ingredients, follow verbal commands, and manipulate utensils, even in settings it has never seen (e.g., through few/zero-shot learning).

Nevertheless, applying M3T-FMs to embodied AI introduces \textit{new challenges that call for a migration from their conventional centralized training/fine-tuning.} In essence, each robot/agent experiences the world through its own \textit{embodiment}: different sensors, actuators, tasks, and user interactions. 
Further, these robots/agents operate in decentralized physical environments (e.g., homes, hospitals, and factories), where through their embodiment they accumulate rich, contextual, and privacy-sensitive data (e.g., confidential industrial processes) that \textit{cannot} be easily pooled/centralized at scale. 
Subsequently, to truly realize the potential of M3T-FMs in these settings, we should move toward \textit{cross-embodiment learning}, where embodied agents that are inherently data collectors can share, refine, and adapt M3T-FMs through decentralized collaborations despite their heterogeneous configurations.
 Here, Federated Learning (FL) offers a compelling mechanism for this collaboration, enabling distributed agents to share model updates without transmitting raw data, thereby preserving privacy~\cite{xianjia2021federated,savazzi2021opportunities}.

In this work, we introduce M3T \textit{Federated Foundation Models} (FFMs) for embodied AI, a natural yet underexplored solution to the challenges and motivations outlined above. Integration of M3T-FFMs in this domain creates a new paradigm that brings together the expressive generalization power of M3T-FMs with the privacy-preserving, decentralized adaptation/learning capabilities of FL. To give our discussions a unified theme, we identify the most relevant aspects of embodied AI that affect the implementation of M3T-FFMs over the  network edge under
 \textbf{EMBODY} dimensions: \underline{\textbf{E}}mbodiment heterogeneity (hardware, sensors, actuators),
\underline{\textbf{M}}odality richness and imbalance,
\underline{\textbf{B}}andwidth and compute constraints,
\underline{\textbf{O}}n-device continual learning,
\underline{\textbf{D}}istributed control and autonomy,
\underline{\textbf{Y}}ielding safety, privacy, and personalization.
This work is created with the purpose of being a \textit{vision paper} that both illuminates the transformative potential of integrating M3T-FFMs in embodied AI and expose the key challenges arising from such integration. Our contributions are summarized below.

\begin{itemize}[leftmargin=3.85mm]
    \item We propose an architecture for M3T-FFMs suitable for embodied AI, featuring modular sensor encoders, Mixture-of-Experts (MoE) layers, and task heads. 
    \item We concretize the \textbf{EMBODY} dimensions, showcase the unique capabilities of M3T-FFMs for embodied AI,  and highlight various use cases of M3T-FFMs in this domain.
    \item We outline various research directions grounded in the \textbf{EMBODY} dimensions, unveiling the unique opportunities that the modularity of M3T-FFMs offer for embodied AI. 
   These directions are intentionally framed to capture the broader theme of \textit{``what can be done"} in this underexplored research area, offering a flexible conceptual basis for future decomposition into specific, actionable research studies.
    \item We envision an evaluation framework for M3T-FFMs in embodied AI, consisting of different evaluation metrics and tradeoffs.     
\end{itemize}

\begin{table*}[t]
    \centering
    \caption{Comparative analysis of FL, M3T-FMs, and M3T-FFM approaches in the embodied AI domain.}
    \vspace{-3mm}
    \includegraphics[width=\linewidth]{table_1.jpg}
    \label{tab:fl_fm_ffm_comparison}
    \vspace{-5mm}
\end{table*}

\begin{figure*}[!h]
    \centering
    \includegraphics[width=\textwidth]{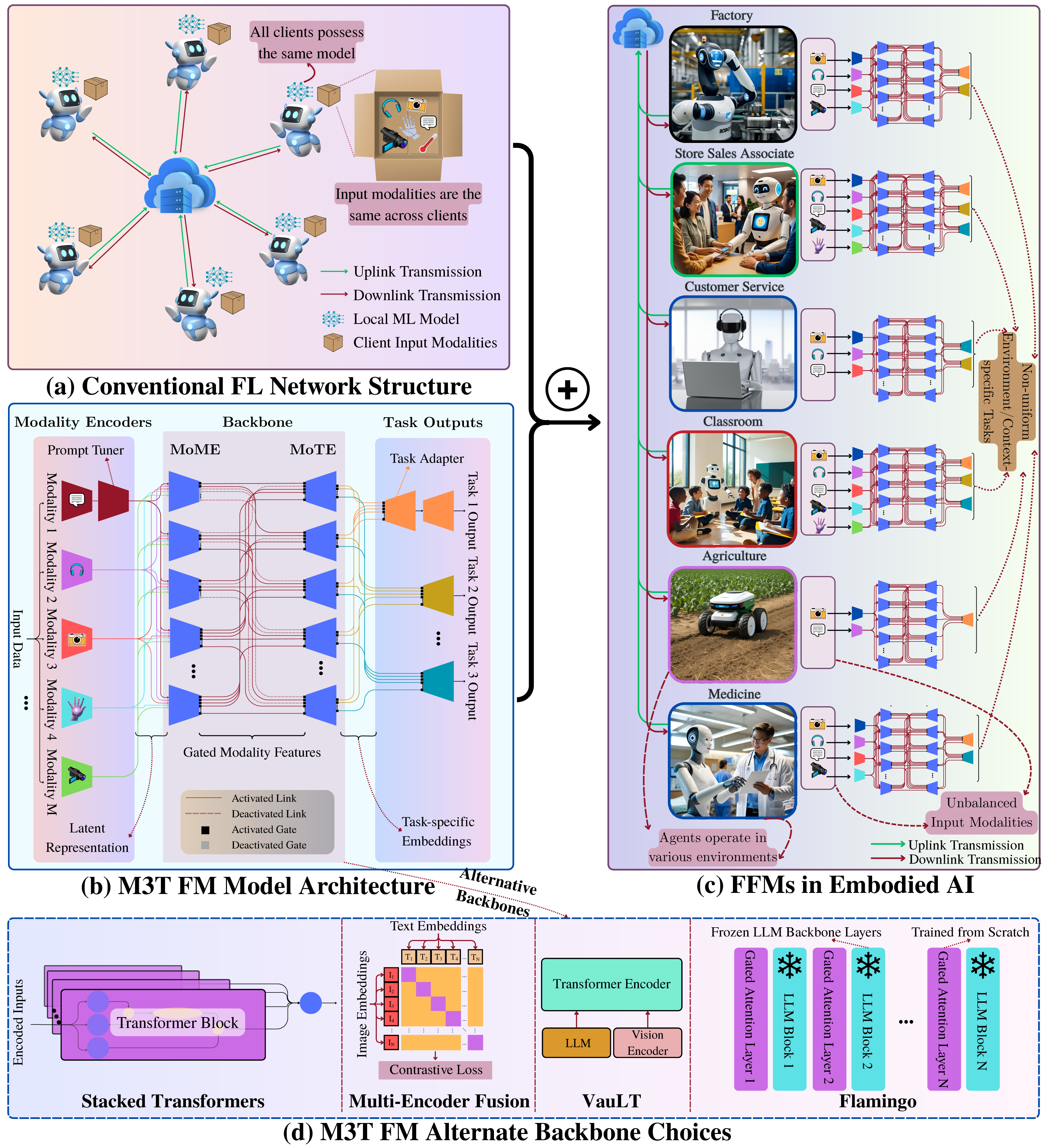}
    \vspace{-5mm}
    \caption{\textbf{(a):} A schematic of FL architecture, where clients/agents engage in collaborative model training. The collaborative training occurs through the repetition of (i) local model training at the agents using their local data, (ii) model/gradient transmission to the server, and (iii) updating the global model at the server based on the received models/gradients (e.g., via weighted averaging) and the broadcast of the global model from the server to the agents to initiate the next round of local model training. \textbf{(b):} Modular architecture of M3T-FMs, comprising modality encoders, MoMEs, MoTEs, and task heads. 
    Based on the input characteristics, a subset of MoMEs are triggered/engaged in the inference and training. Further, based on the desired output tasks, a subset of MoTEs will be activated.    
    \textbf{(c)}: Architecture of FFMs in embodied AI, where different agents have different modalities and tasks of interest. Each agent possesses a local FM, trains different modules (e.g., encoder, task head, subset of MoMEs or MoTEs) of its local FM, and transmits them to the server for aggregation. The server aggregates the received modules and broadcasts these aggregated modules back to the agents. The received aggregated modules can further go through a local fine-tuning at the agents. \textbf{(d)}: Alternative model backbone structures in FMs comprising of stacked transformers (e.g., ChatGPT), multi-encoder fusion (e.g., CLIP), VauLT, and Flamingo (with frozen pre-trained LLM blocks and gated attention blocks trained from scratch).}
    \label{fig:system_figure}
\end{figure*}

\section{Background and Related Work}

\subsection{FL in Robotics/Embodied AI}
Conventional FL (see Fig.~\ref{fig:system_figure}(a)) operates through a repeated three-step process until model convergence: (i) each client/device trains a local model using its data; (ii) model updates, such as parameters/gradients, are periodically shared with a server/aggregator; and (iii) the aggregator combines these updates (e.g., via weighted averaging) to a global model and broadcasts it to the devices, thereby synchronizing their local models and initiating the next round of training. 
FL has been applied in robotics and embodied AI for tasks such as cooperative driving and collaborative manufacturing, enabling robots to jointly learn motion plans and safety-critical controls~\cite{xianjia2021federated,savazzi2021opportunities}. These applications particularly benefit from FL’s few-shot learning capabilities and its ability to generalize across diverse environments and robot embodiments~\cite{10776968}.
Collectively, these advances underline FL's role in scalable intelligence for embodied AI; however, these works do not focus on the FMs.

\subsection{FMs for Embodied AI}

FMs have evolved rapidly, starting with single-modal LLMs (e.g., GPT-3), followed by multi-modal FMs, such as DALL-E. More recently, M3T-FMs  (e.g., GPT-4) have emerged, aiming for general-purpose AI that learns/reasons/acts across diverse tasks and modalities. Although FMs are new to embodied AI, some pioneering works exist: RT-1~\cite{brohan2022rt} trained a transformer on robot trajectories; RoboCat~\cite{bousmalis2023robocat} explored cross-embodiment learning; SayCan~\cite{ahn2022can} and ChatGPT-for-Robotics~\cite{vemprala2024chatgpt} applied LLMs for planning and code generation; UniAct~\cite{zheng2025universal} studied FM-driven actions for embodied AI agents; ECBench~\cite{dang2025ecbench} proposed a benchmark to evaluate FM-enabled embodied AI agents. These studies show the promise of FMs in embodied AI but assume \textit{centralized} training and overlook the modularity of modern M3T-FMs, aspects that we  explore in this work.

\subsection{FFMs for Distributed Embodied AI}
FFMs, especially when considering the emerging M3T-FFMs, are a highly recent research topic. Subsequently, they are quite unexplored in embodied AI. Nevertheless, recent research on FFMs in other domains has shown their tremendous potential, and the study of their aggregation methods and computational/communication efficiency is gaining substantial interest~\cite{ren2025advances}.
In this work, we aim to provide one of the first visions for the integration of M3T-FFMs in the embodied AI domain. To provide a  structured understanding, in
Table~\ref{tab:fl_fm_ffm_comparison}, we compare FL, M3T-FMs, and M3T-FFMs in embodied AI.

\subsection{M3T-FMs Modular Architecture and M3T-FFMs Operations}
There is no unified architecture for M3T-FMs as they are still under active development and envisioned differently across tech companies and academic literature.
In this work, we build upon the proposal architecture in~\cite{chen2024disentanglement} and 
 decompose it into a modular architecture, which is depicted in Fig.~\ref{fig:system_figure}(b), where updates/training can be applied to its various modules independently. 
Subsequently, to enhance the comprehension of the M3T-FFMs, we put them into the context of embodied AI, as illustrated in Fig.~\ref{fig:system_figure}(c), and explain their components below.

\textbf{1. Modality Encoders:} Each sensory input (e.g., RGB-D images, audio, force/torque signals, inertial readings) is processed through an encoder to transform raw signals into latent representations. This modular integration of encoders enables modality-specific fine-tuning and inter-agent encoder swapping without touching the shared backbone defined below.

\textbf{2. Shared Backbone:} The backbone consists of a set of mixture-of-experts (MoEs) described below. 

\textbf{~~(i) Mixture-of-Modality Experts (MoMEs):} To account for heterogeneity in sensing capabilities and workload balancing, latent representations of modalities are passed through a MoME layer. Experts, which are neural/transformer networks, are selectively activated based on the input characteristics, enabling efficient specialization without full-model activation.

\textbf{~~(ii) Mixture-of-Task Experts (MoTEs):} To handle the wide range of embodied tasks, such as navigation, object manipulation, gesture following, or environmental interaction, task-specific MoEs are integrated into the model's pipeline. These allow the model to dynamically activate relevant expert pathways based on a task prompt or contextual signal.

Through the above MoEs, which could be initially pretrained (e.g., through the data scraped from public websites) or trained from scratch alongside other modules~\cite{chen2024disentanglement}, the backbone fuses information across modalities and tasks. Also, it captures compositional structure, spatiotemporal relationships, and contextual grounding necessary for embodied AI, while remaining \textit{mostly} (but \textit{not entirely}) frozen during deployment. 

We note that the model backbone can follow more conventional architectures, such as stacked transformers, multi-encoder fusion, VauLT, and Flamingo, which are depicted at the bottom of Fig.~\ref{fig:system_figure}. 

\textbf{3. Task Heads:} Each embodied task is supported by output heads, which are neural layers that map shared features into concrete predictions (e.g., control commands, action probabilities). This modular integration of heads enables task-specific fine-tuning and inter-agent task head swapping.

\textbf{4. Adapters and Prompts:} Small/shallow adapter modules and/or prompt tuners can be inserted into the backbone or prepended to the input. These modules support agent adaptation or embodiment-specific tuning (i.e., adapting to physical/sensory characteristics of a specific embodied agent).

\textbf{5. Coordinator and Learning Process:} A central coordinator/server manages model updates across agents, following the standard FL process (i.e., local training, aggregation, and broadcast) \textit{without exchanging full-model parameters}. Here, agents update only local FM sub-components/modules (e.g., encoders, heads, adapters, expert weights, prompts), whose selection can be optimized as we discuss later in the \textit{future research directions}. The coordinator aggregates these modules and returns them to agents for further updates. Specifically, the aggregation process mirrors that of conventional FL, with the key distinction that it operates at the \textit{module level} rather than the entire model (e.g., through weighted averaging of module parameters or alternative forms of inter-module knowledge sharing~\cite{chen2024disentanglement}). This modular update scheme enables flexible \textit{module coordination} strategies tailored to each module’s role and dynamics. For instance, subsets of expert/MoE modules may undergo asynchronous, low-frequency aggregation if they already exhibit a high performance. Conversely, task heads associated with emerging or rapidly evolving tasks may benefit from frequent, synchronous updates for accelerated convergence.

$\star$ Henceforth, we use `FFM' to refer to `M3T-FFM' since our focus is solely on this type.



\begin{figure*}[!h]
    \centering
    \includegraphics[width=\textwidth]{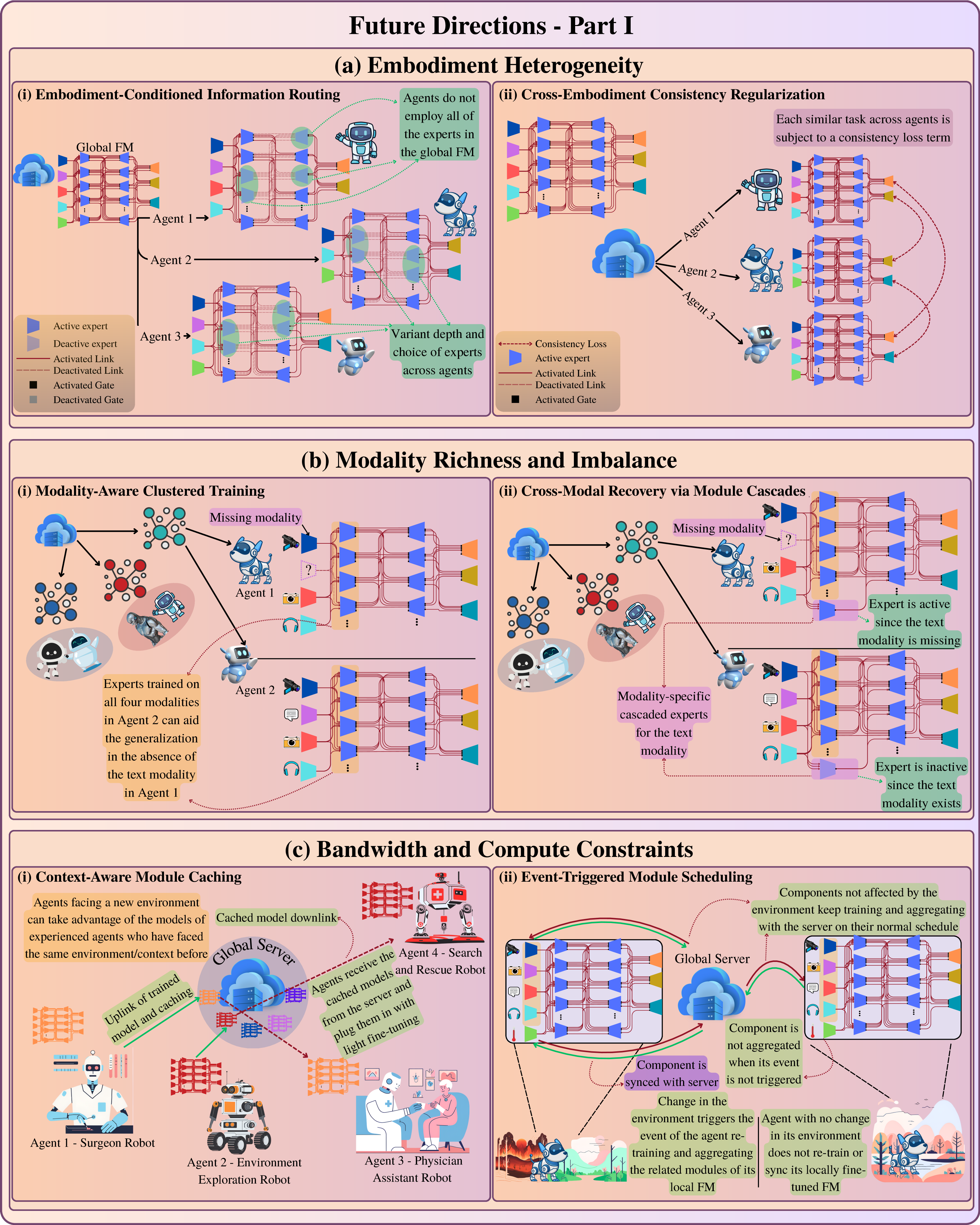}
    \vspace{-5mm}
    \caption{Visualizations of the envisioned future research directions. \textbf{(a):} Embodiment Heterogeneity. Left Plot: Embodiment-Conditioned Information Routing. Right Plot: Cross-Embodiment Consistency Regularization.
    \textbf{(b):} Modality Richness and Imbalance. Left Plot: Modality-Aware Clustered Training.
    Right Plot: Cross-Modal Recovery via Module Cascades. 
    \textbf{(c):} Bandwidth and Compute Constraints.   Left Plot: Context-Aware Module Caching.
    Right Plot: Event-Triggered Module Scheduling.}
    \label{fig:future_directions}
\end{figure*}

\section{Use Cases of FFMs in Embodied AI and \textbf{EMBODY} Dimensions}
Given the aforementioned notable success of FMs in the embodied AI, FFMs can take this one step further and transform a broad spectrum of embodied AI applications across industrial, domestic, and immersive environments.
In the following, we provide some examples and then unveil the natural presence of \textbf{EMBODY} dimensions while articulating them.

\subsection{Use Cases of FFMs in Embodied AI} We next provide three examples on the use cases of FFMs in embodied AI:
\begin{enumerate}[leftmargin=4mm]
\item \textit{Smart Factories:} FFMs can empower cooperative robots with diverse sensors, actuators, and policies to adapt to dynamic workflows and human collaborators. A typical workflow involves robotic perception and data gathering (e.g., via vision or tactile inputs), analysis and policy refinement using task‑specific modules (e.g., using a task head that outputs predictive maintenance schedules), and execution of coordinated actions in real-time.
Performance requirements in such an environment include sub-second response time, low task failure rate, and high sample efficiency for adapting to new tasks (i.e., fast adaptations using few data samples).


\item \textit{Domestic Environments:} FFMs enable assistive robots to learn and personalize to users' routines, preferences, and spaces. 
A typical workflow may involve preparing morning coffee, cleaning while the user is away, restocking items when supplies run low, and reminding of evening tasks, while adapting each action to evolving household habits through on-device continual learning.
Key performance targets in such settings include low computation footprint of model updates (i.e., light adaptations), continual learning without catastrophic forgetting, and high user satisfaction with minimal supervision.
    

\item \textit{Immersive XR Systems:} FFMs allow distributed headsets and wearables to collaboratively fine-tune models for body-language, gaze, and gesture recognition. 
A typical workflow may involve capturing hand gestures, gaze, and speech during an XR session, rendering real-time feedback such as object manipulation or contextual overlays, and periodically syncing refined modality encoder modules across devices to improve recognition accuracy.
Critical performance metrics include sub-second inference latency, high gesture recognition accuracy, and seamless adaptation to new users and environments within a few interactions.

\end{enumerate}

$\star$ Unlike traditional FMs that often require centralized retraining, FFMs support decentralized training and adaptation in all the above scenarios, thereby preserving proprietary/user-specific data and complying with various privacy regulations. Further, different from conventional FMs, which are frozen after centralized pretraining and often lack situational responsiveness, FFMs unlock continuous, privacy-preserving adaptation to new users, tasks, and hardware configurations in all the aforementioned scenarios.

\subsection{Manifestation of \textbf{EMBODY} Dimensions}
In the above scenarios, the \textbf{EMBODY} dimensions naturally emerge, underscoring the critical challenges that naive implementations of FFMs in embodied AI cannot address.

\begin{itemize}[leftmargin=3.85mm]
    \item \textit{\underline{\textbf{E}}mbodiment heterogeneity} spans all the aforementioned scenarios: in smart factories, robots differ in morphology, sensing, and actuation; in homes, assistive robots must adapt to user-specific layouts and hardware configurations; and in XR systems, headsets and embodied avatars vary in tracking precision, interface latency, and sensor fidelity. 
    \item \textit{\underline{\textbf{M}}odality richness and imbalance} is also central across scenarios: factory robots are exposed to vision, force feedback, and machine states; domestic agents interpret multi-modal cues like voice, touch, and gaze; and XR systems may rely on the fusion of head pose, hand motion, eye-tracking, and speech.
    \item \textit{\underline{\textbf{B}}andwidth and compute constraints} are common across scenarios: factory robots operate with limited communication windows and rely on on-board processors that cannot support large-scale model retraining; domestic robots are often low-power, cost-sensitive devices lacking powerful GPUs for full-model fine-tuning; XR systems demand ultra-low latency and high frame rates, precluding heavy model updates or large-scale communication during the execution of XR tasks.
    \item \textit{\underline{\textbf{O}}n-device continual learning} manifests itself universally: robotic arms in factories must adapt to new tasks or tools, home assistants to changing user routines, and XR avatars to evolving behavioral signals, all requiring (near) real-time local model/behavior updates without server interventions.  
    \item \textit{\underline{\textbf{D}}istributed control and autonomy} is intrinsic to these scenarios: factory robots may need to coordinate actions without frequent centralized commands, household robots often operate semi-independently across rooms or homes, and XR users can move independently in immersive environments.   
    \item \textit{\underline{\textbf{Y}}ielding safety, privacy, and personalization} is a shared imperative across the scenarios: safety/regulatory compliance in factories, user privacy in domestic settings, and individualized experiences in immersive systems all demand personalized intelligence that respects privacy and operates within safety margins. 
\end{itemize}



\section{Open Research Directions: An \textbf{EMBODY}-Aligned Agenda for FFMs}
We next revisit the aforementioned \textbf{EMBODY} dimensions, aiming to tailor a series of open research directions (ORD). \textit{\textbf{The directions are intentionally framed at a high-level to allow for diverse interpretations and encourage innovation in this underexplored domain.}}


\begin{figure*}[!h]
    \centering
    \includegraphics[width=\textwidth]{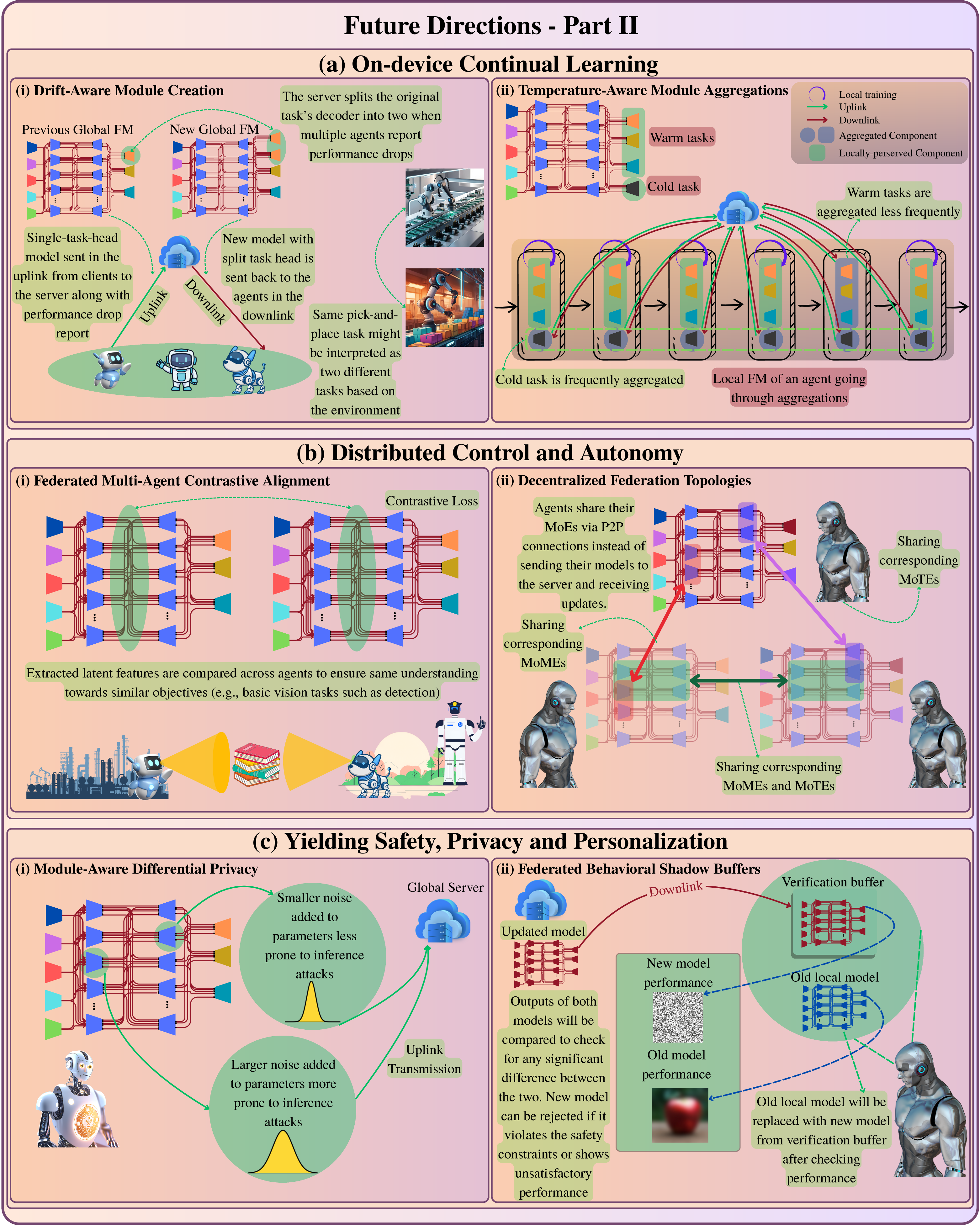}
    \vspace{-5mm}
    \caption{Visualizations of the envisioned future research directions. 
    \textbf{(a):} On-Device Continual Learning.    Left Plot: Drift-Aware Module Creation.
    Right Plot: Temperature-Aware Module Aggregations.
    \textbf{(b):} Distributed Control and Autonomy.  Left Plot: Federated Multi-Agent Contrastive Alignment.
    Right Plot: Decentralized Federation Topologies.
    \textbf{(c):} Yielding Safety, Privacy, and Personalization.  Left Plot: Module-Aware Differential Privacy.
    Right Plot: Federated Behavioral Shadowing Buffers.}
    \label{fig:future_directions2}
\end{figure*}

\subsection{\textbf{\underline{E}}mbodiment Heterogeneity}\label{sec:A}
The embodied heterogeneity, caused by agents operating in various physical environments, poses a challenge to building shared FFMs that generalize across embodiments (e.g., can be trained on one or more types of robots and still perform effectively when deployed on others). To address this, we pose the following overarching ORD (see Fig.~\ref{fig:future_directions}(a)):

\textit{\underline{ORD 1:}  Embodiment-Aware Information Alignment Mechanisms:}
We envision modules at the early layers of the model (e.g., at the MoME layers), which are augmented with locally learned tokens that encode hardware traits (e.g., kinematic range, actuator count, sensor precision). These tokens can then act as soft keys for module activations (e.g., expert selection and activation), enabling intelligent information routing decisions inside the model that control the flow of data processing and adapt it to the physical context during inference and model personalization. This approach, which we refer to as \textit{Embodiment-Conditioned Information Routing}, addresses embodiment heterogeneity by enabling fine-grained, context-aware model specialization across agents with diverse morphologies without requiring hardcoded rules or full-model retraining. As a complementary approach, to avoid agents with different embodiments from diverging in representation space (i.e., variations in hardware, sensing, or control may lead to different internal encodings of similar tasks), we envision the design of consistency loss functions between related tasks performed by heterogeneous embodiments, which can be conceptualized as \textit{Cross-Embodiment Consistency Regularization}, to make a closer connection between similar tasks performed across different agents.

\subsection{\textbf{\underline{M}}odality Richness and Imbalance}\label{sec:B} 
To handle modality richness and imbalance of agents, caused by the (temporal) variations of their sensory inputs, we pose the following ORD (see Fig.~\ref{fig:future_directions}(b)):

\textbf{\textit{\underline{ORD 2:} Cross-Agent Modality Compensation Strategies:}}
We envision clustering agents with similar tasks while ensuring that the combined modalities within each cluster span a broader spectrum than any single agent can provide. Within these clusters, the server can perform federated aggregation across selected modules (e.g., MoMEs), enabling agents to benefit from peers with shared objectives as well as those that expand their sensing capabilities. This approach, which we refer to as \textit{Modality-Aware Clustered Training}, allows modules within an agent’s model (e.g., MoMEs) to uncover hidden structures and relationships between modalities by leveraging knowledge from agents with richer modality sets.
As an additional strategy, to address missing or degraded data streams across agents, we envision \textit{Cross-Modal Recovery via Module Cascades}, where a series of specialized modules (e.g., within the MoME layer) activate only when certain modalities are unavailable or unreliable. These specialized modules should be trained in a federated manner to allow knowledge sharing across these modules from different agents, which increases their generalizability. 
This way, agents can collaboratively learn to fill in their missing modalities (e.g., predicting touch input from vision and movement sensors).

\subsection{\textbf{\underline{B}}andwidth and Compute Constraints}\label{sec:C}
To consider the fact that many embodied agents operate on edge processors with strict energy/compute and bandwidth limitations, we pose the following ORD (see Fig.~\ref{fig:future_directions}(c)): 

\textbf{\textit{\underline{ORD 3:} Resource-Aware Module Usage and Update:}} 
We envision resource‑aware edge caching for FFMs, where \textit{servers} (ranging from cloudlets and edge servers in mobile edge computing deployments, eNodeB base stations with local computing capabilities in cellular networks, road‑side units in vehicular edge computing scenarios, to gateway nodes in generic fog computing architectures) cache/store the modules received from agents during model aggregations, while adhering to their memory constraints. Agents can query their nearest server for modules (e.g., MoEs or adapters) relevant to their current task and environment (e.g., ``following the user in a hallway"), enabling module download and reuse with only lightweight fine‑tuning. This approach, which we refer to as \textit{Context-Aware  Module Caching}, when accompanied by resource allocation strategies (e.g., bandwidth allocation and uplink/downlink transmit power control), 
can enable resource-efficient module usage in bandwidth‑ and memory‑limited wireless edge environments.
As an auxiliary method, we envision \textit{Event‑Triggered Module Scheduling}, where agents selectively update or train specific local modules (e.g., encoders, adapters, or MoEs) \textit{only when} substantial task shifts or performance degradations occur. For instance, a mobile robot may refresh its navigation task head only when entering a new environment or encountering unfamiliar obstacles to conserve communication/computation resources.

\subsection{\textbf{\underline{O}}n-Device Continual Learning}\label{sec:D}

To handle on-device continual learning of embodied agents, caused by facing evolving environments, user preferences, and task definitions --- see Fig.~\ref{fig:future_directions2}(a) --- we pose the following ORD:

\textbf{\textit{\underline{ORD 4:} Temporally-Tuned Module Creation and Aggregation:}}
We envision managing \textit{task drift} in FFMs, where changes in task characteristics over time may require adjusting the model architecture. At the agent level, significant task drift may trigger the fine‑tuning or splitting of a local task head into more specialized variants (e.g., a home robot’s generic ``cleaning" task head evolving into distinct ``kitchen cleaning" and ``bedroom cleaning" heads). This approach, which we refer to as \textit{Drift‑Aware Module Creation}, enables the creation of new task heads at the global model when drift patterns are common across multiple agents, while isolating the task head creation only to an agent's local model when the drift is local.
As an accompanying approach, we envision \textit{Temperature‑Aware Module Aggregation}, where the server adjusts the aggregation weight and frequency of a module based on the module’s temperature, reflecting its real-time readiness or maturity. Well‑trained stable modules (i.e., ``warm" modules) may be frozen and aggregated less often to reduce communication overhead, while newer or low-performing ones (i.e., ``cold" modules) can be aggregated more frequently to accelerate their convergence.

\begin{table*}[t]
    \centering
    \caption{EMBODY dimensions and their related performance metrics (Part 1); Key trade-offs in FMM design for embodied AI (Part 2); Benchmarking FFMs with existing embodied AI datasets (Part 3).}
    \vspace{-3mm}
    \includegraphics[width=\linewidth]{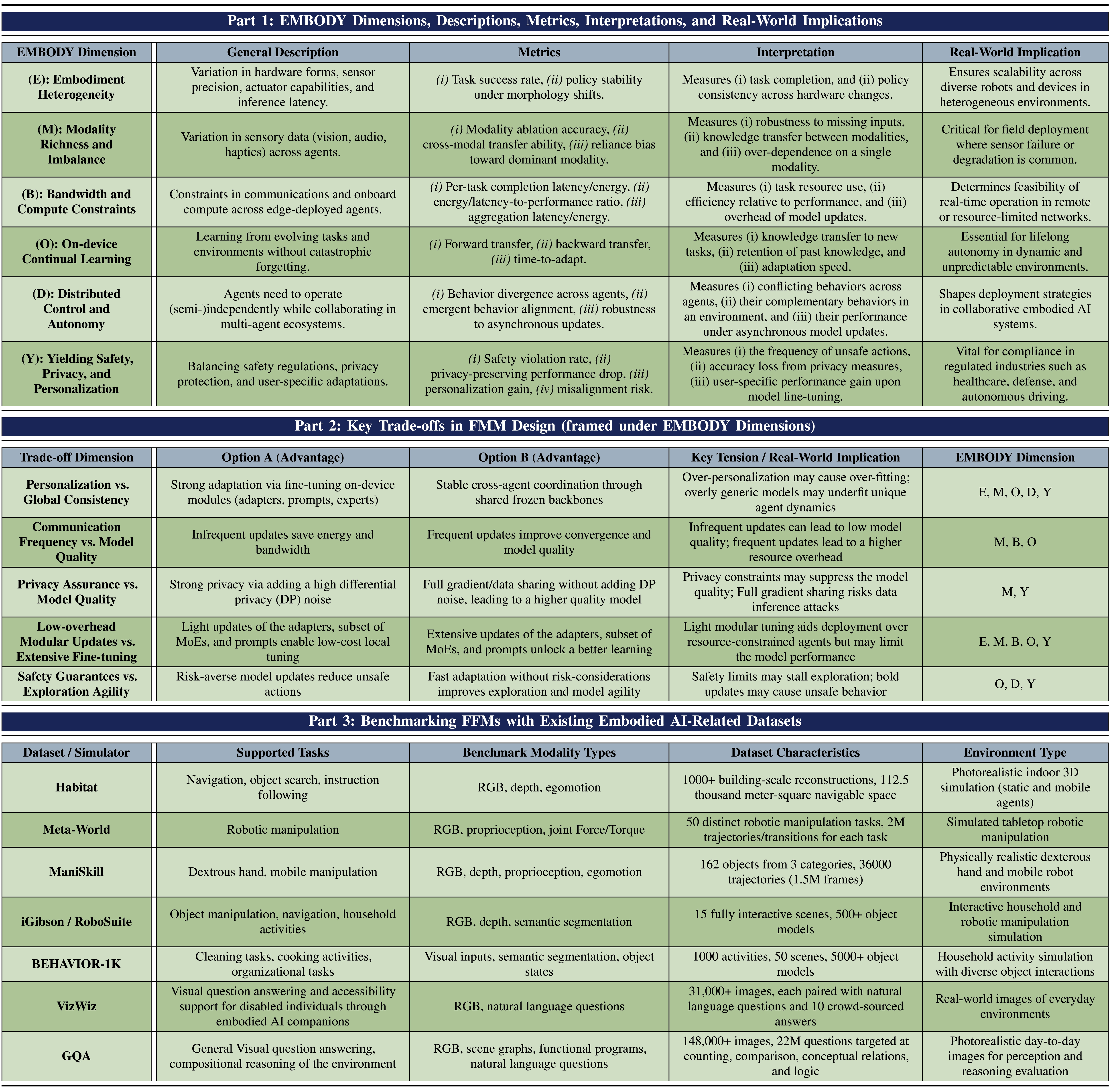}
    \label{tab:ffm_tradeoffs_and_benchmarks}
    \vspace{-5mm}
\end{table*}

\subsection{\textbf{\underline{D}}istributed Control and Autonomy}\label{sec:E} 

To enable embodied agents to adapt to their local tasks and contexts in a coordinated yet decentralized manner --- see Fig.~\ref{fig:future_directions2}(b) --- we pose the following ORD:

\textbf{\textit{\underline{ORD 5:} Advanced Cross-Agent Model Refinement Techniques:}}
We introduce the concept of cross-agent refinement of FFMs through complementary strategies at both the \textit{learning} and \textit{network} levels. At the learning level, we envision \textit{Federated Multi‑Agent Contrastive Alignment} with the ultimate goal of aligning the latent task representations of agents that perform similar tasks under different environments, hardware configurations, or user preferences. This can be achieved by using a contrastive loss objective to pull together internal representations of similar tasks and push apart those of unrelated tasks across the agents (e.g., aligning the obstacle‑avoidance task representations of an aerial drone and a ground robot).
At the network level, we envision \textit{Decentralized Federation Topologies}, where agents exchange their modules (e.g., MoMEs, MoTMs, adapters, encoders, or task heads) via low-power, short-range peer‑to‑peer (P2P) communication to reduce the reliance on resource-intensive uplink transmissions to the central server. 
Depending on agent proximity and communication topology, this can follow \textit{fully decentralized} deployments, where all model exchanges occur through P2P links without any uplink usage (suitable for small‑scale, fully connected networks), or \textit{semi‑decentralized} deployments, where local P2P exchanges are combined with occasional uplink transmissions from selected agents to merge knowledge across distant agents. Notably, such P2P module exchanges can be guided by mutual‑trust evaluation mechanisms between agents, a topic of long‑standing research \cite{10547257}.






\subsection{\textbf{\underline{Y}}ielding Safety, Privacy, and Personalization}  
To make agents comply with safety, privacy, and user-specific constraints --- see Fig.~\ref{fig:future_directions2}(c) --- we pose the following ORD:

\textbf{\textit{\underline{ORD 6:} Techniques for Module-Level Privacy and Fault Resilience:}}
We envision strengthening privacy and operational robustness in FFMs under the presence of malicious entities in the network and/or non-ideal connectivity conditions (e.g., network jitter and packet loss).
At the privacy level, we envision \textit{Module‑Aware Differential Privacy} mechanisms that apply fine‑grained, non‑uniform noise to different modules (e.g., encoders, task heads, or MoME/MoTE experts) before transmission for aggregation to obscure the innate information that is encoded in the model parameters. Specifically, the noise intensity must be tuned based on each module’s characteristics; for example, modality encoders that process sensitive visual or audio streams would receive an amplified injected noise to avoid the possibility of sensitive information recovery. This can be further complemented by module‑level functional encryption (e.g., homomorphic encryption), enabling aggregation of encrypted module parameters without revealing their raw values.
At the fault‑tolerance level, we propose \textit{Federated Behavioral Shadow Buffers}, where model updates from the server, which might be distorted by network jitter and packet loss, are first deployed in a ``shadow mode" that generates predictions without enacting actions. Agents compare these predictions to those from prior models, monitor deviations, and incorporate user feedback or corrections before applying the updates to their local models. This validation can also occur in a virtual environment (e.g., digital twin) to assess model outputs without real‑world risks.

\begin{figure*}
    \centering
    \includegraphics[width=1\linewidth]{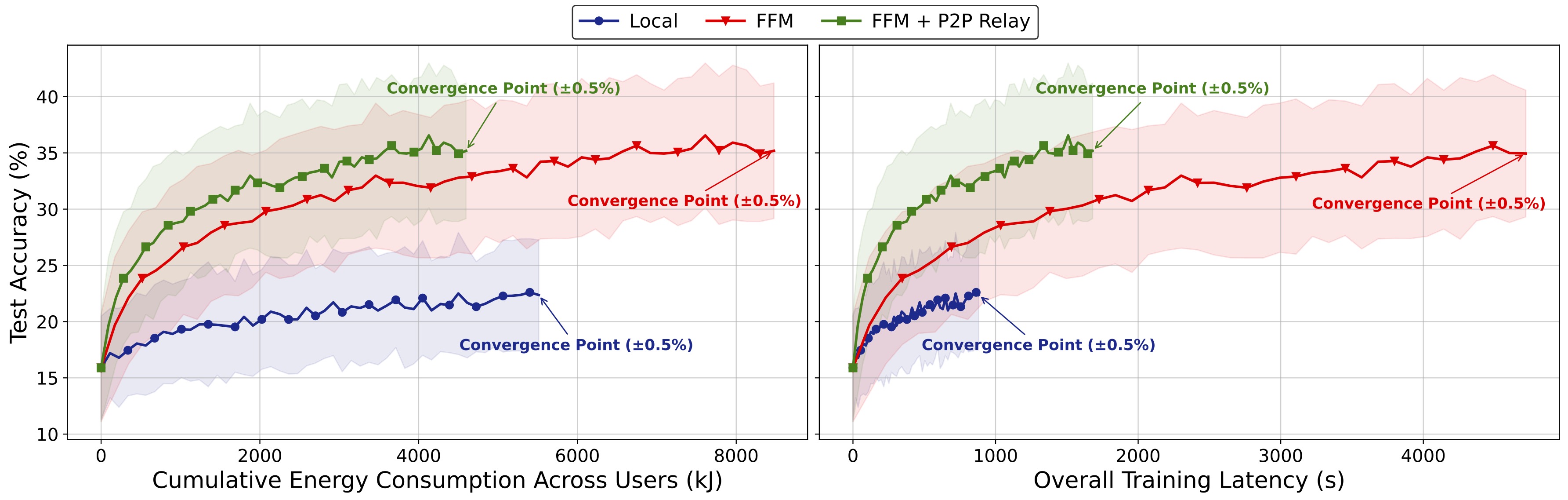}
    \vspace{-7mm}
    \caption{Test accuracy (averaged over the two tasks) versus cumulative energy consumption across users (left subplot)  and overall training latency (right subplot) for different methods. The ``Convergence Point ({\scriptsize $\pm0.5\%$})" markers denote the energy level and latency at which each method converges and oscillates with only less than {\scriptsize $\pm0.5\%$} in accuracy. FFM + P2P Relay consistently delivers faster convergence while converging to the same final accuracy as FFM, underscoring the role of P2P communication among embodied AI agents in minimizing energy usage and latency. Isolated local training, where embodied AI agents do not engage in cross-embodiment/user knowledge sharing, converges to significantly lower accuracy.}
    \label{fig:energy_latency}
    \vspace{-3mm}
\end{figure*}


\section{Evaluation Dimensions of FFMs: An \textbf{EMBODY}-Aligned Perspective}
We next envision an evaluation framework developed around the \textbf{EMBODY} dimensions, structured into three key components as outlined in the overarching Table~\ref{tab:ffm_tradeoffs_and_benchmarks}.
In particular, in Part~1 of this table, we present the six \textbf{EMBODY} dimensions and their brief descriptions, followed by the corresponding performance metrics that can be used for their evaluation. For each dimension, we also provide an interpretation of the reported metrics, explaining what they reveal about system behavior and robustness. Further, we discuss the real-world implications of addressing each \textbf{EMBODY} dimension. 
Next, we consider the fact that to operationalize FFMs in embodied AI, it is critical to understand the multi-dimensional trade-offs that are rooted in the evaluation of \textbf{EMBODY} dimensions.  To facilitate comprehension, in Part 2 of Table~\ref{tab:ffm_tradeoffs_and_benchmarks}, we present a structured view of these trade-offs. Each row highlights a tension between two competing design choices, such as personalization vs. global generalization, and articulates their real-world implications. This table can serve as a guide for system designers to tailor FFM architectures and protocols according to the operational priorities and constraints of their target environments.
Finally, in Part 3 of Table~\ref{tab:ffm_tradeoffs_and_benchmarks}, we provide a structured description of key embodied AI datasets that can be used to evaluate FFMs across diverse tasks and modalities.

\section{Prototype of FFMs}
We consider an edge network comprising $35$ embodied AI robots. The edge robots are partitioned into $7$ clusters, where clusters represent different zones/areas ($5$ robots are considered in each cluster, which can further form P2P networks as in ORD 5). 
We presume a scenario where the embodied AI robots aim to learn two heterogeneous tasks/datasets: (i) learning to engage in question-answering (e.g., for answering queries about everyday environments, such as identifying what objects are present, what actions are happening, or where items are located)  via 
recognizing generic images and the description of objects in them via GQA dataset (\url{https://cs.stanford.edu/people/dorarad/gqa/download.html}), and (ii) recognizing different shopping products and their descriptions (e.g., for assisting individuals/customers, particularly those with visual impairments, with shopping) through VizWiz dataset (\url{http://vizwiz.org/tasks-and-datasets/vqa/}), using two data modalities (text and image).



Our implementations are publicly available on GitHub: {\small \url{https://github.com/payamsiabd/M3T-FFM-EmbodiedAI}}, detailing the P2P network topology and uplink/downlink channel models. 
The non-iid distribution of each dataset/task across agents follows Dirichlet distribution~\cite{savazzi2021opportunities} with concentration parameter $0.5$. We adopt ViLT (with the size of 328 MB) as the backbone of the FMs deployed/trained on robots, which offers two advantages: (i) it employs a lightweight text embedding layer instead of the LLM–based text encoder used in VauLT (Fig.~\ref{fig:system_figure}(d)), and (ii) it is designed for image–text multimodality, which matches the modalities present in our datasets.

We fine‑tune the model using lightweight \textit{adapters} embedded within every transformer layer alongside task‑specific output heads (with the total size of 6MB). Each adapter adopts a bottleneck architecture with a hidden dimension of $256$, comprising a down‑projection layer, a GELU nonlinearity, and an up‑projection layer.

We examine the following methods:
\textit{(i) Local FM Deployment:} Each robot trains its own FM without aggregating/knowledge sharing with other robots.
\textit{(ii) FFM Deployment:} After one epoch of local training, each robot sends its fine-tuned adapter and task head module parameters to the cloud, which conducts a module aggregation using \textit{FedAvg}~\cite{savazzi2021opportunities}.
\textit{(iii) FFM + P2P Module Relaying}: 
In each cluster, a randomly selected robot serves as the cluster head and performs the following operations. After completing local training, all robots in the cluster transmit their fine-tuned parameters to the cluster head via multi‑hop P2P links along the shortest path, where the parameters are aggregated. The cluster head then solely uploads its aggregated parameters to the cloud for global aggregation, thereby reducing dependence on resource‑intensive uplink transmissions.


In Fig.~\ref{fig:energy_latency}, we capture various performance metrics, such as energy-to-performance ratio (left subplot) and latency-to-performance ratio (right subplot), where the performance is measured via the test accuracy. Comparing the performance of the \textit{`Local'} baseline with other FFM variants highlights the role of global knowledge sharing in FFMs, where the local model's performance saturates to a low accuracy. Further, observing the plots reveals the energy usage and latency of FFM fine-tuning processes. 
Specifically, the \textit{FFM + P2P Relay} achieves the same converged accuracy as its \textit{FFM} counterpart, attributable to replicating model aggregations via multi‑hop P2P model exchanges, while incurring lower energy consumption and latency, mainly due to the efficient communications enabled by low‑cost, short‑range P2P links.

\section{Conclusion}
In this paper, we envisioned that by integrating the generalization power of M3T-FMs with the decentralized learning capabilities of FL, M3T-FFMs can offer a unified learning approach in the embodied AI ecosystems. Through the \textbf{EMBODY} framework, we articulated the core dimensions that should be addressed for M3T-FFMs to succeed in real-world embodied AI deployments. For each dimension, we identified open challenges and envisioned actionable research directions that span model architecture, training dynamics, and system configuration. We also discussed evaluation protocols and trade-off analysis for M3T-FFM deployment over embodied AI.  {We highlight that \textbf{EMBODY} provides a structured foundation that can serve as a benchmark framework for the community. By aligning future work with the \textbf{EMBODY} dimensions and their associated evaluation metrics, researchers can systematically compare methods, quantify trade-offs, and ensure that research progress is both measurable and reproducible.}

\bibliographystyle{IEEEtran}
\bibliography{Ref}
\end{document}